\documentclass[10pt, conference]{IEEEtran}
\IEEEoverridecommandlockouts
\usepackage{cite}
\usepackage{amsmath,amssymb,amsfonts}
\usepackage{algorithmic}
\usepackage{graphicx}
\usepackage{textcomp}
\usepackage{xcolor}
\usepackage{enumerate}
\usepackage{comment}
\usepackage{xpatch}
\xpatchcmd\maketitle{\@fnsymbol}{\@arabic}{}{}
\xpatchcmd\maketitle{\rlap}{}{}{}

\def\BibTeX{{\rm B\kern-.05em{\sc i\kern-.025em b}\kern-.08em
    T\kern-.1667em\lower.7ex\hbox{E}\kern-.125emX}}
\newcommand{\MSCOMMENT}[1]{\textcolor{red} {(MS: #1)}}

\title{TEA-DNN: the Quest for Time-Energy-Accuracy Co-optimized Deep Neural Networks}

\author{
\IEEEauthorblockN{Lile Cai*}
\IEEEauthorblockA{I2R, Singapore \\
caill@i2r.a-star.edu.sg}
\and
\IEEEauthorblockN{Anne-Maelle Barneche*}
\IEEEauthorblockA{SUPELEC, France \\
anne-maelle.barneche@supelec.fr}
\and
\IEEEauthorblockN{Arthur Herbout*}
\IEEEauthorblockA{ECP, France \\
arthur.herbout@student.ecp.fr}
\and
\IEEEauthorblockN{Chuan Sheng Foo} 
\IEEEauthorblockA{I2R, Singapore \\
foo\_chuan\_sheng@i2r.a-star.edu.sg}
\and
\IEEEauthorblockN{Jie Lin} 
\IEEEauthorblockA{I2R, Singapore \\
lin-j@i2r.a-star.edu.sg}
\and
\IEEEauthorblockN{Vijay Ramaseshan Chandrasekhar\textsuperscript{\textdagger}}
\IEEEauthorblockA{I2R, Singapore \\
vijay@i2r.a-star.edu.sg}
\and
\IEEEauthorblockN{Mohamed M. Sabry\textsuperscript{\textdagger}}
\IEEEauthorblockA{NTU, Singapore \\
msabry@ntu.edu.sg}
\thanks{\textsuperscript{*}Equal contribution. \textsuperscript{\textdagger}Joint corresponding authors.}
}

\IEEEpubid{\makebox[\columnwidth]{978-1-7281-2954-9/19/\$31.00 $\copyright$2019 IEEE \hfill{ \hspace{\columnsep}\makebox[\columnwidth]{ }}}}
\begin{document}
\maketitle

\begin{abstract}

Embedded deep learning platforms have witnessed two simultaneous improvements. First, the accuracy of convolutional neural networks (CNNs) has been significantly improved through the use of automated neural-architecture search (NAS) algorithms to determine CNN structure. Second, there has been increasing interest in developing hardware accelerators for CNNs that provide improved inference performance and energy consumption compared to GPUs. Such embedded deep learning platforms differ in the amount of compute resources and memory-access bandwidth, which would affect performance and energy consumption of CNNs. It is therefore critical to consider the available hardware resources in the network architecture search. To this end, we introduce TEA-DNN, a NAS algorithm targeting multi-objective optimization of execution time, energy consumption, and classification accuracy of CNN workloads on embedded architectures. TEA-DNN leverages energy and execution time measurements on embedded hardware when exploring the Pareto-optimal curves across accuracy, execution time, and energy consumption and does not require additional effort to model the underlying hardware. We apply TEA-DNN for image classification on actual embedded platforms (NVIDIA Jetson TX2 and Intel Movidius Neural Compute Stick). We highlight the Pareto-optimal operating points that emphasize the necessity to explicitly consider hardware characteristics in the search process. To the best of our knowledge, this is the most comprehensive study of Pareto-optimal models across a range of hardware platforms using actual measurements on hardware to obtain objective values.
\end{abstract}


\section{Introduction}
\label{Intro}
Deep convolutional neural networks (CNNs) have achieved state-of-the-art performance in image classification, object detection and many other applications \cite{goodfellow2016deep}. To achieve better accuracy, CNN models have become increasingly deeper and require more computing and memory resources \cite{he2016identity,szegedy2017inception}. This poses a challenge when these models are deployed to run on resource-limited devices, such as mobile and embedded platforms, as the memory on these devices may not be large enough to hold the models or running the model may consume more power than the device can supply.

Much effort has been devoted to designing CNN models that can run efficiently on these devices, for instance, by manually designing more efficient convolution operations and network architectures \cite{howard2017mobilenets,zhang1707shufflenet,huang2017condensenet}. However, this approach demands expert knowledge and obtaining an optimal model is difficult -- one has to carefully balance the trade-off between accuracy and computational resources. An alternative approach is to use automated neural architecture search (NAS) algorithms to find optimal models under hardware constraints \cite{zoph2016neural,zoph2017learning,liu2017progressive}. NAS algorithms usually consist of three components: a controller, a trainer and an evaluator. The controller is responsible for sampling models from the search space. The trainer is then responsible for training the sampled models. Finally, the evaluator is responsible for evaluating the optimization objectives (e.g., model accuracy) on currently sampled models. Following this evaluation, the parameters of the controller are updated to increase the likelihood that it subsequently samples better models. 

\begin{figure}[h]
	\centering
	{\includegraphics[width=0.9\linewidth]{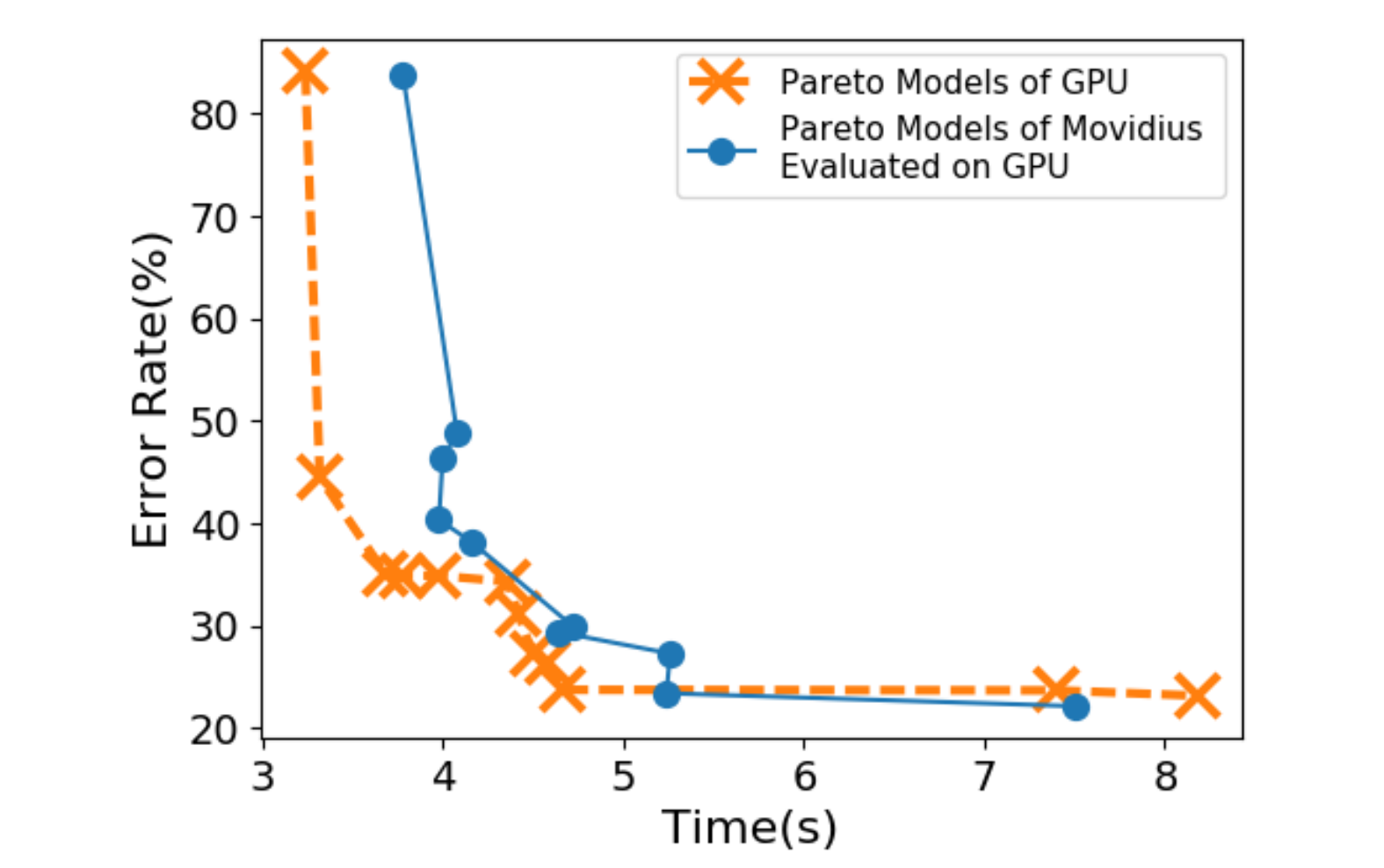}}
    \vspace{-3mm}
	\caption{Pareto-optimal models searched on Movidius are suboptimal on TITAN X GPU.}
    \label{killer_pic}
\end{figure}

Due to the variation in platform hardware/software configurations, models optimized for one platform can be suboptimal for another. Consider two hardware platforms, namely the Nvidia TITAN X GPU~\cite{titanx_gpu} and Intel Movidius Neural Computing Stick (NCS)~\cite{movidius} that respectively exemplify a high-performance and embedded platform. Figure~\ref{killer_pic} displays two Pareto-optimal curves (inference time versus classification error) of CNN models targeting TITAN X GPU and NCS, both executed on the TITAN X GPU (for measurement details see Section~\ref{experiments}). It can be seen that the Pareto-optimal models searched for the NCS are far from the Pareto curve for the GPU, implying that a platform-agnostic NAS may result in highly suboptimal models -- in this case resulting in up to $2\times$ increase in execution time to achieve comparable accuracy.


The problem revealed in Fig.~\ref{killer_pic} demonstrates that in order to obtain an optimal model for a hardware platform, its corresponding characteristics have to be taken into consideration during the search process. To this end, we introduce TEA-DNN (Time-Energy-Accuracy co-optimized Deep Neural Networks), a NAS framework that explicitly considers two hardware metrics -- inference time and energy consumption -- in addition to classification accuracy as objective metrics. We formulate the neural architecture search problem as a multi-objective optimization problem and leverage Bayesian optimization to search for Pareto-optimal solutions. While Bayesian optimization has been used to obtain hardware-aware neural networks \cite{stamoulis2018hyperpower}, it was only used to search for several hyper-parameters with a fixed network architecture. To the best of our knowledge, our work is the first to apply Bayesian optimization for neural architecture search. Furthermore, TEA-DNN does not require modeling the hardware platform and instead leverages the ability to directly measure energy and execution time on actual hardware. We summarize our contributions as follows:

\begin{itemize}
\item A time, energy and accuracy co-optimization framework for CNNs. 

\item Employing Bayesian optimization to search for CNN structures that yield Pareto-optimal operating conditions.

\item We demonstrate how different device configurations can lead to different trade-off behaviors.

\item We demonstrate that optimal models searched on one hardware platform are not optimal for another and thus reiterate the importance of hardware-aware NAS.
\end{itemize}


\section{Related Work}
\label{related_work}
\subsection{Neural Architecture Search (NAS)}
Early versions of NAS algorithms~\cite{zoph2016neural} employed recurrent neural networks (RNNs) to predict the architecture of a target CNN where the weights of the RNN are updated using reinforcement learning. \cite{zoph2017learning} follows the same framework as proposed in \cite{zoph2016neural}, but instead of using a RNN to predict the entire network architecture, the algorithm only predicts the optimal structure for one convolutional module (or ``cell''). Identical cells are then stacked multiple times to form the full network. \cite{liu2017progressive} replaced reinforcement learning with progressive search, which can yield better models with fewer samples.

\subsection{Hardware-Aware NAS}
Explicitly incorporating hardware constraints into NAS has been an active research topic in recent years. HyperPower \cite{stamoulis2018hyperpower} approximates the power and memory consumption of a networks using linear regression. These approximations are then used in the acquisition function of a Bayesian optimization algorithm to avoid sampling models that violate power or memory constraints. MnasNet \cite{tan2018mnasnet} focuses on finding optimal networks for mobile devices and used inference time as one of the objectives. DPP-Net \cite{dong2018dpp} performs neural architecture search on different devices and considers more optimization objectives: error rate, number of parameters, FLOPs, memory, and inference time.

While our work is closely related to MnasNet and DPP-Net in that we all search for Pareto-optimal networks for a specific device, our approach is unique in two respects. Firstly, we perform true multi-objective optimization instead of combining several objectives into a single objective as done in MnasNet. Secondly, unlike DPP-Net, we do not use any surrogate functions to approximate the optimization objectives. Instead, we directly measure the \emph{real-world} values for all the three objectives (i.e., time, energy and accuracy). This eliminates the need to model the targeted hardware, which is a challenging task given the diversity of hardware platform configurations.

\section{TEA-DNN Optimization Framework}
\label{method}
\subsection{System Overview}
We formulate the neural network architecture search problem as a multi-objective optimization problem $\boldsymbol{\min}_x (error(x), energy(x), time(x))$
where we wish to find a network architecture parameterized by $x$ (see Section \ref{sec_ss} for details) that minimizes classification \emph{error}, \emph{energy} consumption, and inference \emph{time}. We do not assume a closed-form model for energy consumption or inference time, but evaluate them directly on actual hardware to measure real-world performance. Networks were trained and evaluated on GPUs for efficiency as we assume that classification error is not affected by the specific hardware a network is run on. 

As formulated, this is an instance of a black-box optimization problem where the objective functions can only be evaluated (and are not differentiable), and where function evaluations (especially classification error, which requires training the model) are costly. Note that no single ``best solution'' exists for a multi-objective optimization problem. A solution is instead defined by a Pareto optimal set of points, for which improvement in any objective function cannot be made without negatively affecting some other objectives.

\begin{figure}[htbp]
	\centering
	{\includegraphics[width=\linewidth]{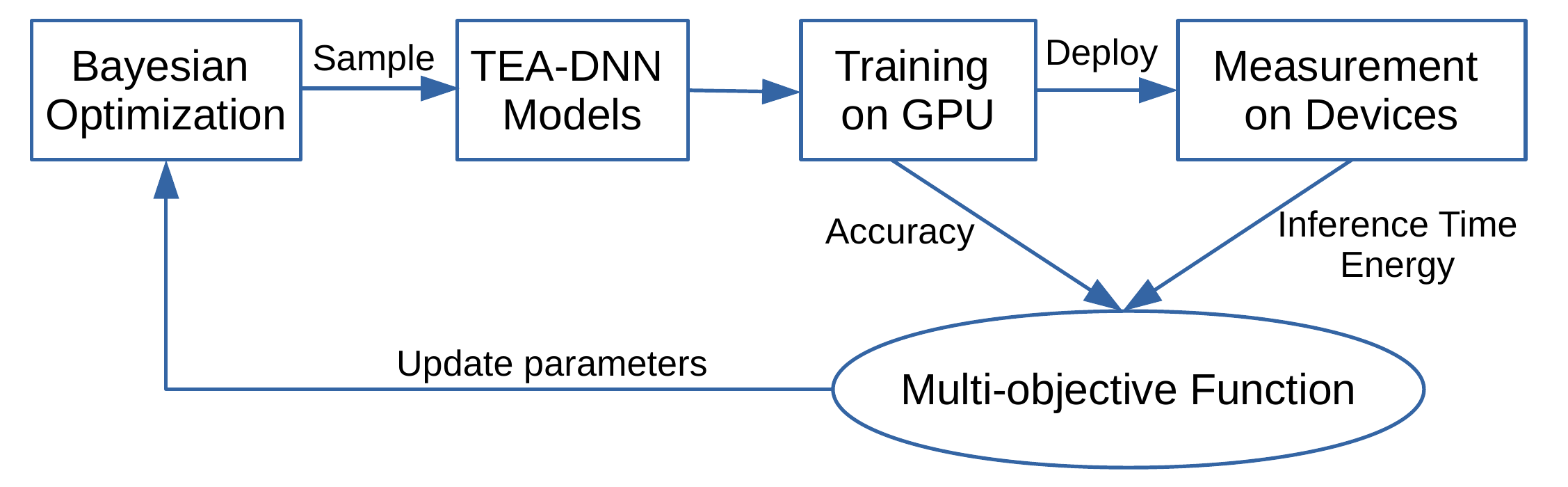}}
    \vspace{-8mm}
	\caption{System diagram of the proposed TEA-DNN optimization framework.}
     \label{system_diagram}
\end{figure}

We chose to employ a Bayesian optimization algorithm \cite{hernandez2016predictive} (detailed in Section \ref{sec_mobo}) to solve this optimization problem. We provide a brief overview and refer the reader to the comprehensive review in \cite{shahriari2016taking}. Bayesian optimization algorithms perform a sequential exploration of the parameter space while building a surrogate probabilistic model to approximate the objective functions. This model is used to \emph{select points} at which to next \emph{evaluate} the objective functions, and the obtained function values are then used to \emph{update the model}. The algorithm proceeds iteratively following this \emph{select-evaluate-update} loop, such that points in the Pareto optimal set are selected more frequently as the algorithm progresses. We stopped the algorithm after 400 points are sampled. A schematic overview of our search algorithm is shown in Fig.~\ref{system_diagram}.

\subsection{Search Space}
\label{sec_ss}

\begin{figure}[h]
	\centering
	{\includegraphics[width=\linewidth]{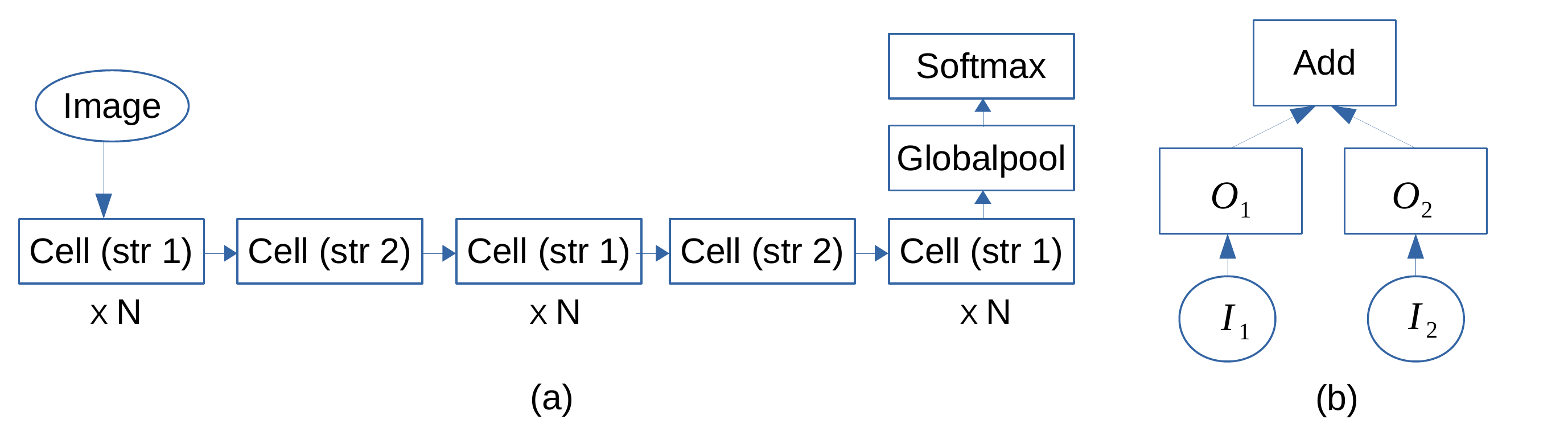}}
    \vspace{-7mm}
	\caption{The network architecture predefined for CIFAR-10 (a) and the architecture of one building block (b).}
	\label{building_block_cell}
\end{figure}

We search over the subset of network architectures that can be described as repetitions of a modular network ``cell'', as proposed by \cite{liu2017progressive}. The overall network architecture is predefined (illustrated in Fig.~\ref{building_block_cell}(a)) and consists of cells with either stride 1 or 2 (i.e., slide the filter every 1 or 2 pixels). As a common heuristic, the number of filter channels is doubled after the stride 2 cells. As such, the network architecture is uniquely determined by the initial filter channel number $F$, the number of cell repeats $N$ and the cell structure. $F$ and $N$ are hyper-parameters that are pre-specified and the cell structure is searched using Bayesian optimization.

Specifically, each cell is composed of 5 building blocks and each building block (illustrated in Fig.~\ref{building_block_cell}(b)) is parameterized by 4 parameters $(I_1, I_2, O_1, O_2)$ for a 20-dimensional parameter space. $I_1$ and $I_2$ denote the inputs, and $O_1$ and $O_2$ specify the operations applied to the respective inputs. The input space of each building block consists of the outputs of all preceding blocks in the current cell as well as outputs from the two preceding cells. The operation space includes the following eight functions commonly used in top performing CNNs:
\begin{enumerate}
	\item $\mbox{max}\  3\times 3$: $3\times 3$ max pooling
	\item $\mbox{identity}$:  identity mapping
	\item $\mbox{sep}\  3\times 3$: $3\times 3$ depthwise-separable convolution
	\item $\mbox{conv}\  3\times 3$: $3\times 3$  convolution
	\item $\mbox{sep}\  5\times 5$: $5\times 5$ depthwise-separable convolution
	\item $\mbox{conv}\  5\times 5$: $5\times 5$  convolution
	\item $\mbox{sep}\  7\times 7$:  $7\times 7$ depthwise-separable convolution
	\item $\mbox{conv}\  7\times 7$:  $7\times 7$ convolution
\end{enumerate}
The search space described above has an order of $10^{14}$ ($2^2\times 8^2\times 3^2\times 8^2\times 4^2\times 8^2\times 5^2\times 8^2\times 6^2\times 8^2= 5.6\times 10^{14}$). The outputs of the two operations are then combined by element-wise addition. The final output of the cell is the concatenation of all unused building block outputs. 

\subsection{Multi-Objective Bayesian Optimization}
\label{sec_mobo}
Bayesian optimization is a sequential model-based approach that approximates each objective function with a Gaussian process (GP) model. For a particular objective function (e.g., classification error)  let $f(x)$ be its surrogate GP model, $x_{1:n}$ be the evaluated network architectures in the search space, $f_i = f(x_i)$ be the objective function value for network $x_i$, and $y_i$ be the actual measured function value. The GP model assumes that $\mathbf{f} = f_{1:n}$ are jointly Gaussian with mean $\mathbf{m}$ and covariance $\mathbf{K}$ and observations $\mathbf{y} = y_{1:n}$ are normally distributed given $\mathbf{f}$:
\begin{equation}
\begin{array}{l}
\mathbf{f} \sim N(\mathbf{m}, \mathbf{K}),\\
\mathbf{y}|\mathbf{f},\sigma^2 \sim N(\mathbf{f}, \sigma^2\mathbf{I}).
\end{array}
\end{equation}

Each iteration of Bayesian optimization consists of 3 steps:
\begin{enumerate}
\item Selecting the next point (network architecture to evaluate) $x_{n+1}$ by maximizing an acquisition function, which specifies a likely candidate that improves the objective(s). We used the PESMO (Predictive Entropy Search Multi-objective) \cite{hernandez2016predictive} acquisition function in our experiments that chooses points which maximally reduce the entropy of the current posterior distribution given by the GPs over the Pareto set.

\item Evaluating the objective functions at $x_{n+1}$.
\item Updating the parameters $\mathbf{m}$ and $\mathbf{K}$ for the GP models.
\end{enumerate}

To employ Bayesian optimization for neural architecture search, we use the 20-dimensional parameterization of the search space as described in Section~\ref{sec_ss}. Our three objective functions are the 1) error rate (i.e., $1-\mbox{accuracy}$), 2) inference time and 3) energy consumption, and we used the open-source PESMO implementation in Spearmint\cite{hernandez2016predictive} for our experiments.

\section{Experimental Setup}
\label{experiments}
We evaluate TEA-DNN models on different deep-learning hardware platforms, representing embedded and server-based systems. Table~\ref{device_spec} summarizes the properties of these platforms.
\begin{table}

\caption{Specifications of the devices used in our experiments.}
\begin{tabular}[width=\linewidth]{|r|c|c|c|} \hline
                          &  GTX TITAN X  & Jetson TX2      & Movidius  \\ \hline          
  Processing Unit         &  3072 CUDA cores     & 256 CUDA cores  & Myriad 2 VPU    \\ \hline
  FLOPS                   &  6.7T FP32         & 1.5T FP32        & 2T FP16     \\ \hline
  Memory                  &  12GByte GDDR5      & 8GByte LPDDR4     & 4GBit LPDDR3     \\ \hline
  Mem. Bandwidth        &  336.6 GBytes/s          & 59.7 GBytes/s       & 4 GBits/s \\ \hline
  Power                   &  250 W               & 15 W           &  1 W\\ \hline
  
\end{tabular}
\label{device_spec} 
\end{table}

\subsection{Training Setup in Search Process}
In our experiments, models are trained and tested on the CIFAR-10 dataset\cite{krizhevsky2009learning}, which is a popular benchmarking dataset for image classification. CIFAR-10 has 50,000 training images and 10,000 test images of dimension $32\times 32 \times 3$. We removed 5,000 images from the training set for use as a validation set and train on the remaining 45,000 images. During the search process, each model is trained for 10 epochs with a batch size of 32. We use the RMSProp optimizer \cite{tieleman2012lecture} with momentum and decay both set to 0.9. The learning rate is set to 0.01, and decayed by 0.94 every 2 epochs. Weight decay is set to 0.00004. For data augmentation, images are
first zero-padded with 4 pixels on each side to $40\times 40$, and then randomly cropped to $32\times 32$, followed by random horizontal flip and random adjustment of brightness and contrast. The initial channel number $F$ is set to 24 and the number of cell repeats $N$ is set to 2 in the search process.

\subsection{Time and Energy Measurement}
\begin{figure}[h]
	\centering
	{\includegraphics[width=\linewidth]{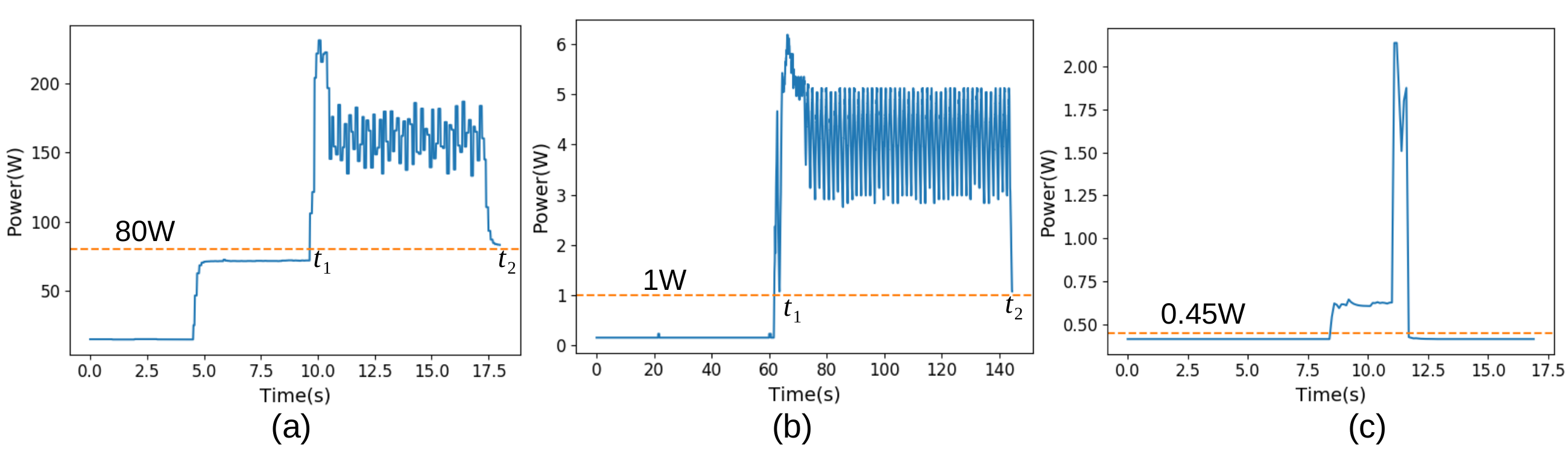}}
	\vspace{-7mm}
	\caption{Power curves on TITAN X GPU (a), Jetson TX2 (b) and Movidius NCS (c). The dashed orange line represents the threshold above which the device is considered under load.}
	\label{power_curves}
\end{figure}

The measurement method for each device is detailed as below:
\begin{itemize}
    \item \textbf{TITAN X GPU}
          A model is launched to run on CIFAR-10 validation subset (5000 images) with a batch size of 100. During the process, power is queried every 20ms using NVIDIA Management Library (NVML) \cite{nvml}. A typical power curve is displayed in Fig.~\ref{power_curves}(a). We use a threshold of 80W to separate the curve into working and idle states. The threshold is decided empirically and we have found that it works well for all models we tested. The inference time is then computed as $t_2 - t_1$ and energy is computed by integration.
    \item \textbf{Jetson TX2}
          A model is launched to run on CIFAR-10 validation subset with a batch size of 100. During the process, power and the corresponding time stamp are obtained using the Python library provided by \cite{Convenient2018}. A typical  power curve is presented in Fig.~\ref{power_curves}(b). The threshold is set to 1W. Inference time and energy are computed similarly as done for TITAN X GPU.
    \item \textbf{Movidius NCS}
          We use the profiling tool provided by the Movidius Neural Compute SDK \cite{ncsdk} to obtain the inference time. During profiling, a power meter \cite{powerz} is attached to the NCS to monitor power consumption. A typical power curve recorded by the power meter is shown in Fig.~\ref{power_curves}(c). The threshold is set to 0.45W  and energy is computed similarly as done for TITAN X GPU and Jetson TX2. 
\end{itemize}

\section{Results and Discussions}

\label{results_discussions}
\subsection{Evolution of Pareto Curves}
\begin{figure*}[h]
	\centering
	{\includegraphics[width=\linewidth]{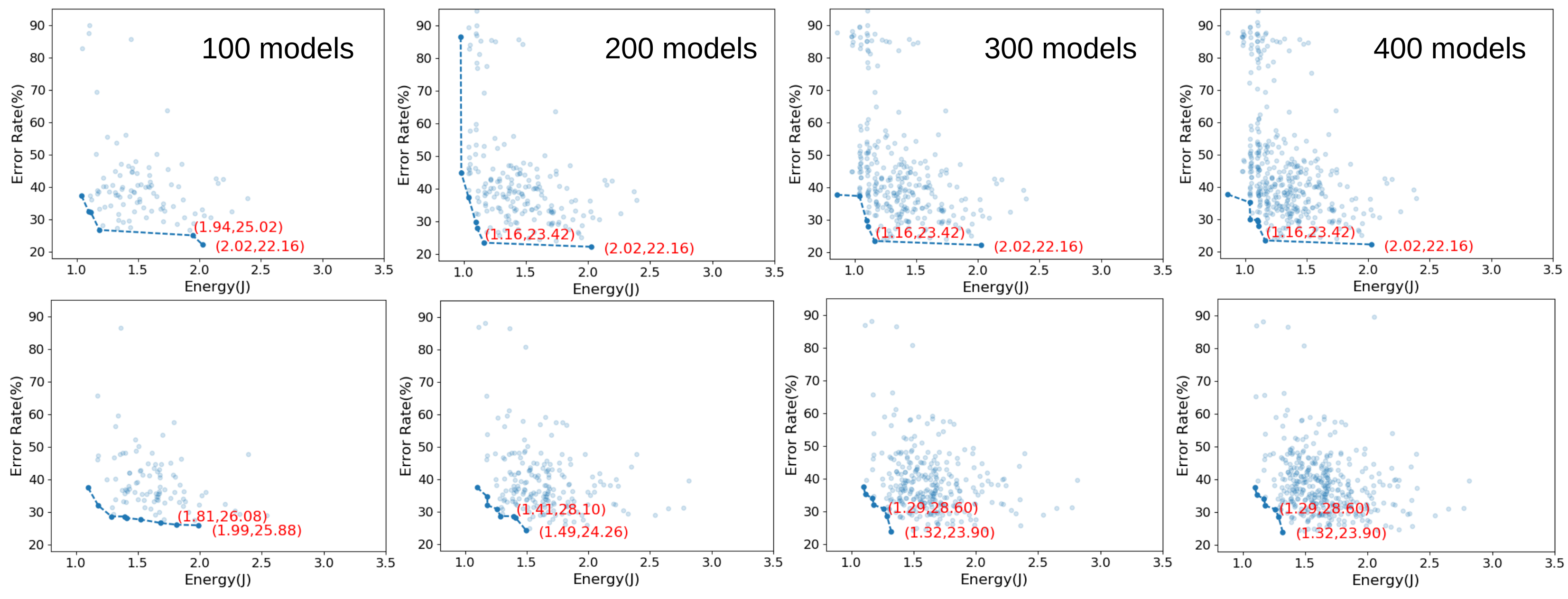}}
    \vspace{-7mm}
	\caption{Evolution of Pareto curves (error rate vs. energy) found by Bayesian optimization (top row) and random sampling (bottom row) on Movidius NCS. Figures are plotted for every 100 sampled models. In each subfigure, we show the coordinates for the last two points for easier comparison.}
     \label{pareto_evolution_movidius_random_energy}
\end{figure*}

To validate the effectiveness of Bayesian optimization in searching time-energy-accuracy co-optimized DNN models, we compare the Pareto curves found by Bayesian optimization to those from random sampling of network architectures on Movidius NCS in Fig.~\ref{pareto_evolution_movidius_random_energy}. Randomly sampled models are generated by selecting inputs and operations for a building block (Section~\ref{sec_ss}) uniformly at random. We observe that Bayesian optimization is able to explore the search space more effectively in that it has a more spread-out distribution of points (models), and more points seem to lie in the Pareto optimal set. Also, it is able to return better models using fewer function evaluations (sampled models). For instance, after sampling 100 models, Bayesian optimization is able to find a model with an error rate of 22.16\% and energy consumption of 2.02J, whereas random sampling can only find a model with much higher error rate (25.88\%) at similar energy levels (1.99J); after sampling 300 models, Bayesian optimization finds models that achieve a better trade-off between error rate and energy: a model with an error rate of 23.42\% and energy consumption of 1.16J, while the best model found through random sampling has a higher error rate (23.90\%) and consumes more energy (1.32J). This clearly demonstrates the effectiveness of Bayesian optimization in searching TEA-DNN models.

\subsection{Cross-Device Evaluation of Pareto-Optimal Models}
\begin{figure}[!h]
	\centering
	{\includegraphics[width=\linewidth]{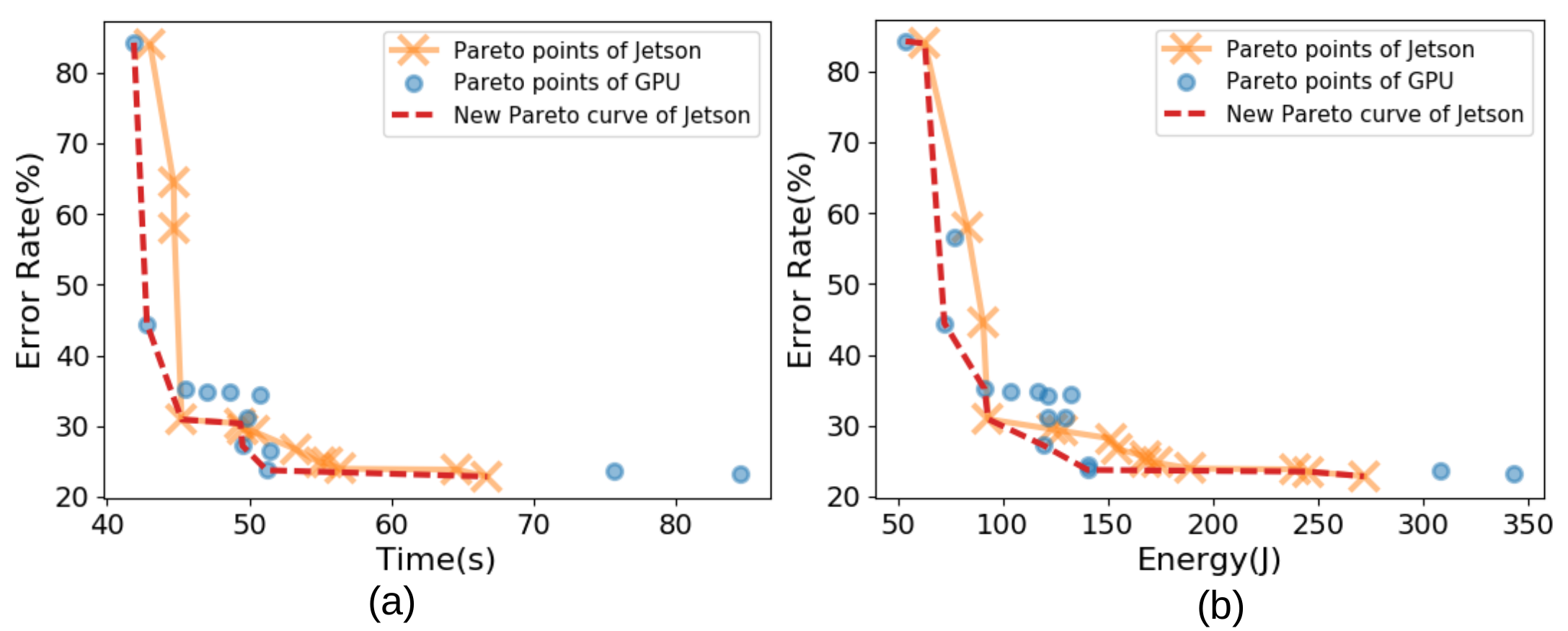}}
	\vspace{-7mm}
	\caption{Evaluating Pareto-optimal models searched for TITAN X GPU on Jetson TX2.}
	\label{gpu_on_jetson}
\end{figure}

\begin{figure}[!h]
	\centering
	{\includegraphics[width=\linewidth]{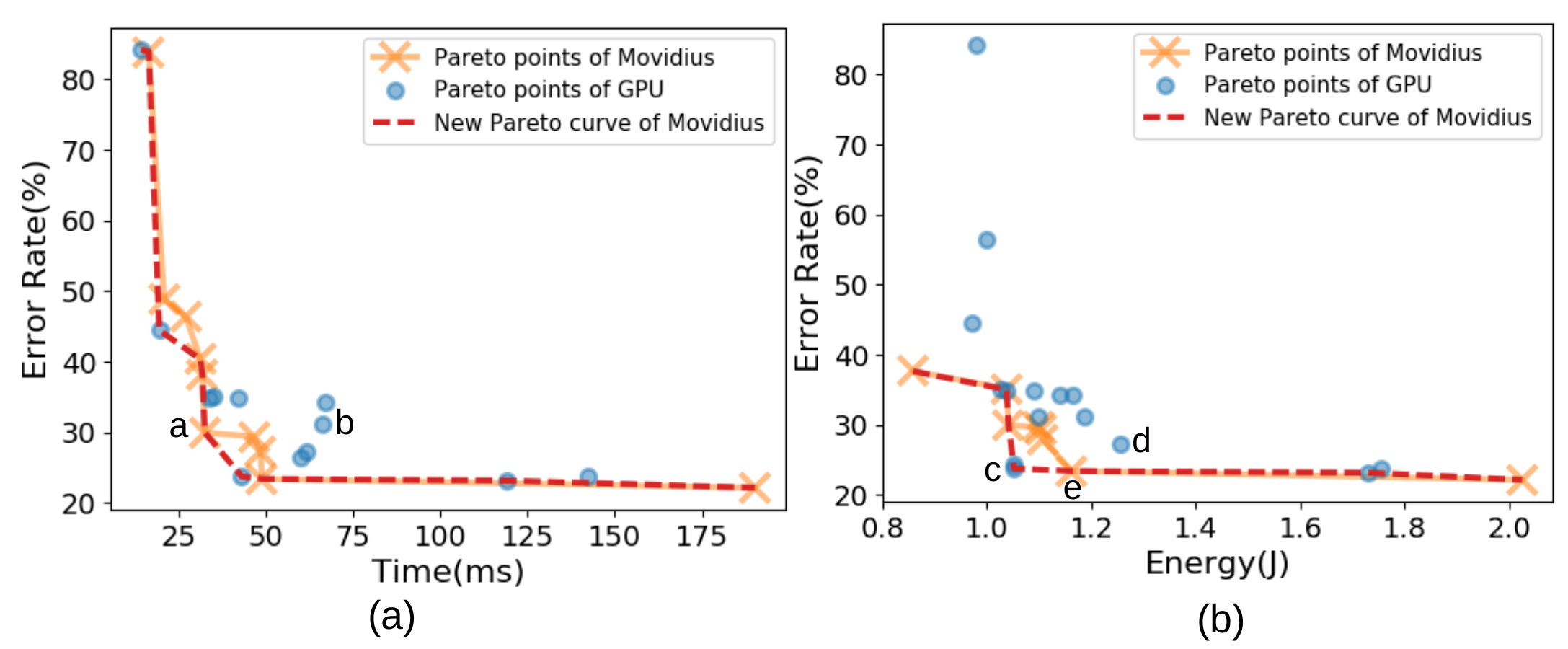}}
	\vspace{-7mm}
	\caption{Evaluating Pareto-optimal models searched for TITAN X GPU on Movidius NCS.}
	\label{gpu_on_movidius}
\end{figure}

When performing neural network search on different platforms, a natural question arises as to whether the set of Pareto-optimal models searched for one platform is also Pareto-optimal for another. We perform cross-device evaluation of Pareto-optimal models to answer this question. First, we evaluate Pareto-optimal models searched for the TITAN X GPU on the Jetson TX2 (Fig.~\ref{gpu_on_jetson})) and Movidius NCS (Fig.~\ref{gpu_on_movidius}). We observe that models optimal on the TITAN X GPU are not guaranteed to be optimal on embedded devices, in that more than half of the Pareto-optimal models for the GPU (blue points) lie to the upper-right (i.e., are inferior in one or both dimensions) of the original Pareto curve for each of the embedded devices (orange line). A GPU-optimal model can incur significantly higher computational cost than a device-optimal model at similar accuracy levels: in Fig.~\ref{gpu_on_movidius}(a), the GPU-optimal model (point b) takes $2\times$ the inference time compared to the device-optimal model (point a) (66.30ms vs. 32.52ms).  We also note that models can behave very differently on different platforms. For example, in Fig.~\ref{gpu_on_movidius}(b), model c consumes more energy than model d on the TITAN X GPU (508J vs. 489J), but less on the Movidius (1.05J vs. 1.26J), and forms the new Pareto curve on Movidius. These highly platform-dependent behaviors clearly indicates that incorporating energy and execution time of the targeted platform is key in TEA-DNN. 

\begin{figure}[!h]
	\centering
	{\includegraphics[width=\linewidth]{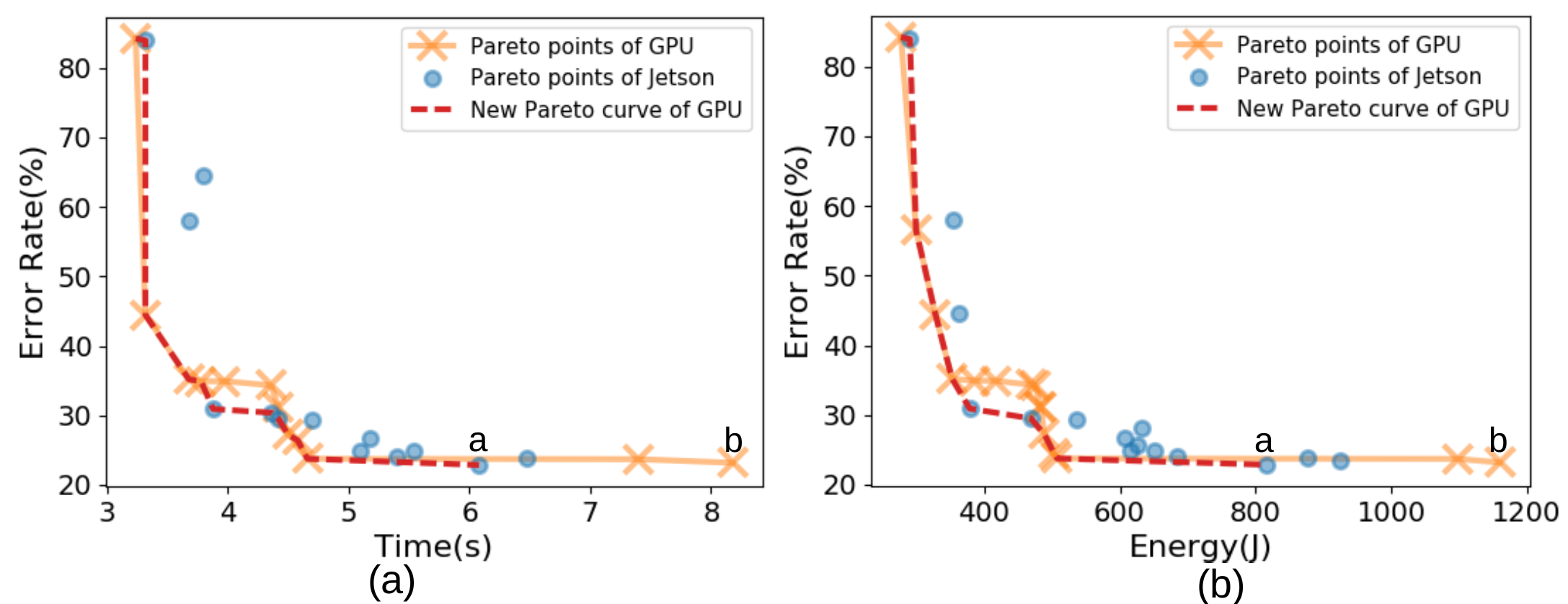}}
	\vspace{-7mm}
	\caption{Evaluating the Pareto-optimal models searched for Jetson TX2 on TITAN X GPU.}
	\label{jetson_on_gpu}
\end{figure}

\begin{figure}[!h]
	\centering
	{\includegraphics[width=\linewidth]{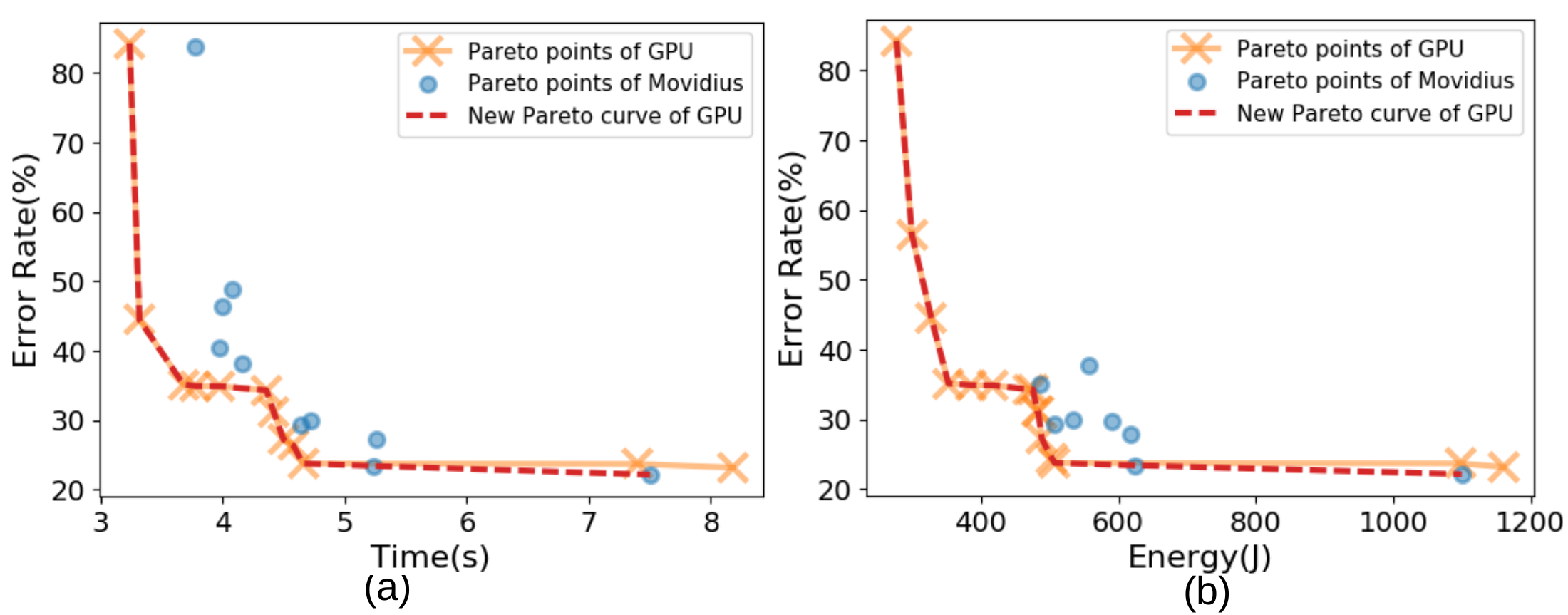}}
	\vspace{-7mm}
	\caption{Evaluating the Pareto-optimal models searched for Movidius NCS on TITAN X GPU.}
	\label{movidius_on_gpu}
\end{figure}

In addition, we evaluate the set of Pareto-optimal models for the Jetson TX2 and Movidius NCS on the TITAN X GPU, as illustrated in Fig.~\ref{jetson_on_gpu} and Fig.~\ref{movidius_on_gpu}. One interesting phenomenon is that the last model on the Pareto curve of the embedded devices can always form the new Pareto curve of GPU. For example, in Fig.~\ref{jetson_on_gpu}(a), the Jetson-optimal model a achieves lower error rate than GPU-optimal model b (22.86\% vs. 23.18\%), but runs faster and less energy than model b (6.08s vs. 8.18s and 815J vs. 1160J), and thus replaces b to become the new Pareto point. This suggests that the limited resources on embedded platforms yield CNNs architectures that consume less compute resources in the high-performance GPU. 

\subsection{Accuracy Benchmarking}
\begin{figure}[!h]
	\centering
	{\includegraphics[width=\linewidth]{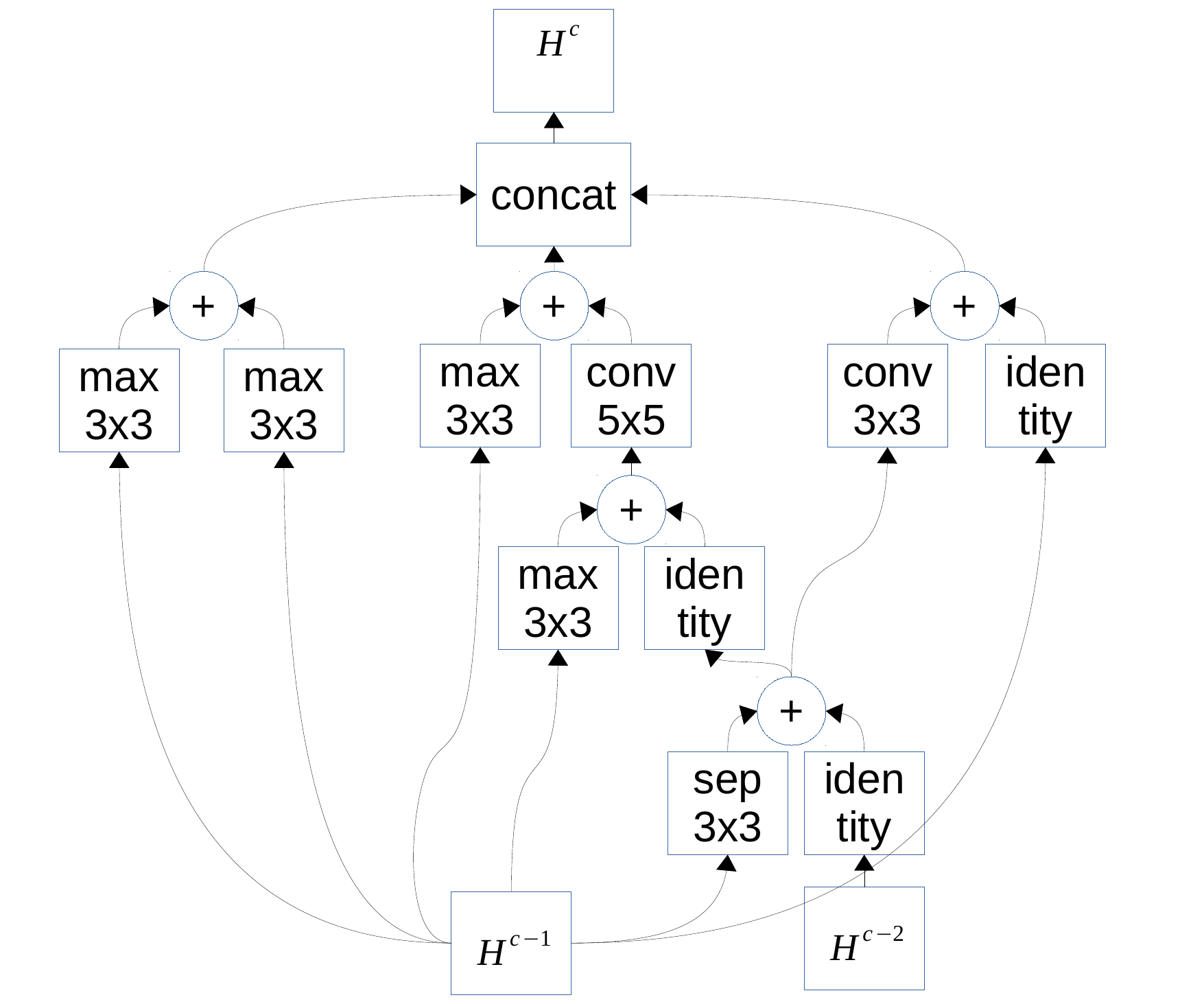}}
    \vspace{-7mm}
	\caption{The cell structure of the TEA-DNN models evaluated in Table~\ref{accuracy_benchmarking}.}
    \label{optimal_structure_time_error_movidius}
\end{figure}

We select a model from the Pareto curve of Movidius that achieves low error rate while does not consume too much energy,  i.e., Model e in Fig.~\ref{gpu_on_movidius}(b). To obtain the final model, we set the batch size to 64 and train for 300 epochs. The initial learning rate is 0.01 and is decayed by 0.5 every 100 epochs. Weight decay is 0.0001. Other settings are the same as the training setup in search process. We experiment with different $N$ and $F$ (Section~\ref{sec_ss}) and the results are reported on the CIFAR-10 test subset (10,000 images) in Table~\ref{accuracy_benchmarking}. It can be seen that deeper (larger N) and wider (larger F) achieves lower error rates, with the cost of more parameters, energy consumption and running time. The cell structure of Model e is shown in Fig.~\ref{optimal_structure_time_error_movidius}. We note that this model employs quite a few parameter-efficient operations, e.g., max-pooling, identity, and $3\times 3$ convolutions.

In Table~\ref{accuracy_benchmarking}, we also compare the TEA-DNN models with state-of-the-art image classification model DenseNet \cite{huang2017densely}. DenseNet is a carefully designed network that connects each layer to every other subsequent layers. It limits the number of feature maps that each layer produces in order to improve parameter efficiency. Such connection pattern and feature map number limit are not employed in TEA-DNN. It can be seen that TEA-DNN model with N=2 and F=12 consumes less energy than DenseNet on TITAN X GPU and Jetson TX2, while DenseNet has lower error rate and faster running time. Note that DenseNet is not currently supported by Movidius NCS as the parser cannot fuse batch norm parameters in the composite function into the concatenated input.

\begin{table}[h]
	\tiny
\caption{Benchmarking of TEA-DNN models on CIFAR-10 test with varying $N$ and $F$.}
\centering
\begin{tabular}{|r|p{3mm}|p{5mm}|p{4mm}p{2mm}p{5mm}|p{4mm}p{2mm}p{5mm}|} \hline
                         & Error & \#Params   & \multicolumn{3}{|c|}{Energy}  & \multicolumn{3}{|c|}{Time}  \\ \hline    
                         &       &        & TITANX & Jetson & Movidius & TITANX & Jetson & Movidius \\\hline
  TEA-DNN (N=2, F=12)   & 10.20  & 0.3M & 690J       & 175J & 1.03J & 7.62s & 94.8s &   54.9ms                  \\ \hline
  TEA-DNN (N=3, F=12)   & 9.79 & 0.5M & 876J       & 233J & 1.09J & 8.38s & 101s &  71.2ms                   \\ \hline 
  TEA-DNN (N=3, F=16)   & 8.22 & 0.9M &  1142J      & 287J & 1.11J & 9.68s & 106s     &  53.3ms         \\ \hline 
  TEA-DNN (N=2, F=24)   & 7.89 & 1.3M &  1131J      & 310J &  1.04J & 9.38s & 107s &  48.9ms         \\ \hline
  TEA-DNN (N=3, F=24)   & 7.44 & 2.0M & 1473J       & 395J & 1.28J & 11.08s & 119s &  66.4ms           \\ \hline
  TEA-DNN (N=3, F=48)   & 6.62 & 8.0M &  2675J      & 730J & 1.73J& 15.76s  & 164s & 121ms           \\ \hline
  DenseNet (L=40, k=12) & 5.44 & 1.1M & 1018J & 207J  & n.a. & 5.24s & 46.5s & n.a.            \\ \hline 
\end{tabular}
\label{accuracy_benchmarking}
\end{table}

\section{Conclusions}
\label{conclusions}
In this work, we propose the TEA-DNN framework that employs Bayesian optimization to search for time-energy-accuracy co-optimized CNN models. We apply TEA-DNN on three different devices: TITAN X GPU, Jetson TX2 and Movidius NCS. Comparison with random sampling shows that Bayesian optimization is able to explore the search space more effectively and to return better models using fewer sampled models. Detailed cross-device evaluation of Pareto-optimal models demonstrates that optimal models searched for one hardware platform are not guaranteed to be optimal for another, and models can behave very differently on different platforms. Our comprehensive experiments reveal the highly platform-dependent behaviours of neural network models and reiterates the importance of explicitly considering hardware characteristics in neural architecture search.



\bibliographystyle{IEEEbib}
\bibliography{references}

\section*{Acknowledgements}
This research is supported by A*STAR under its Hardware-Software Co-optimisation for Deep Learning (Project No.A1892b0026). The computational work for this article was partially performed on resources of the National Supercomputing Centre, Singapore (https://www.nscc.sg). 
\end{document}